\pdfoutput=1

\documentclass[11pt]{article}

\usepackage{acl}

\usepackage{times}
\usepackage{latexsym}

\usepackage[T1]{fontenc}
\usepackage{graphicx}

\usepackage[utf8]{inputenc}

\newcommand{\etal}{\textit{et al.}}
\usepackage{array}
\usepackage{booktabs}
\newcolumntype{F}[1]{%
    >{\raggedright\arraybackslash\hspace{0pt}}p{#1}}%
\newcolumntype{T}[1]{%
    >{\centering\arraybackslash\hspace{0pt}}p{#1}}%

\usepackage{microtype}

\title{EXnet: Efficient In-context Learning for Data-less Text classification}

\author{Debaditya Shome\\
  KIIT University, India\\
  \texttt{1804372@kiit.ac.in} \\\And
  Kuldeep Yadav \\
  SHL, India \\
}

\begin{document}
\maketitle
\begin{abstract}
Large pre-trained language models (PLMs) have made significant progress in encoding world knowledge and spawned a new set of learning paradigms including zero-shot, few-shot, and in-context learning. Many language tasks can be modeled as a set of prompts (for example, is this text about geography?) and language models can provide binary answers, i.e., Yes or No. There is evidence to suggest that the next-word prediction used by many PLMs does not align well with zero-shot paradigms. Therefore, PLMs are fine-tuned as a question-answering system. In-context learning extends zero-shot learning by incorporating prompts and examples, resulting in increased task accuracy.
Our paper presents EXnet, a model specifically designed to perform in-context learning without any limitations on the number of examples. We argue that in-context learning is an effective method to increase task accuracy, and providing examples facilitates cross-task generalization, especially when it comes to text classification tasks. With extensive experiments, we show that even our smallest model (15M parameters) generalizes to several unseen classification tasks and domains.  
\end{abstract}

\section{Introduction}

Large pre-trained language models (LMs) have revolutionized Natural Language Processing (NLP) and demonstrated state-of-the-art results over a varied set of tasks \cite{min2021recent}. Despite the success, most pre-trained models need to be fine-tuned over large corpus of manually-annotated training data, especially when the origin of the data is from a different domain. Annotation of such large-scale task-specific data for training each downstream model is extremely expensive, and a time-consuming process. To solve this issue, recent research focus on zero-shot learning with PLMs that involves solving downstream tasks without any further gradient-based training (parameter updates) of the models. For instance, GPT-3 \citep{brown2020language} \& other GPT-family of models such as GPT-Neo showed an interesting ability of In-context learning in which the model could learn unseen tasks on-the-fly using a few demonstration examples, without any parameter updates.
Zhong \etal \cite{zhong2021adapting} demonstrated that it is possible to convert text classification tasks to Yes/No and a train a prompt-based model to perform zero-shot learning. Unfortunately, pure zero-shot models provide low task-specific accuracy and are not useful for many downstream applications. In-context learning that takes additional context in terms of both prompt and examples outperforms classic Zero-shot models and likely to work better for many real-world applications. We argue that adding a small number of demonstration examples is a small overhead but can help provide a significant accuracy improvements.  Popular PLMs do support In-context learning but have major limitations. Apart from a huge training overhead, they have a significant inference costs and make them inaccessible to a large number of practitioners. Further, they provide poor accuracy on domain-specific tasks, support only limited number of examples, and also, sensitive to the order of examples \citep{zhao2021calibrate}.

In this paper, we propose EXnet, a light-weight model architecture that enables us to utilize In-context support examples for few-shot text classification. Our model doesn't have a limit on the number of input task examples, and there is no visible impact on the ordering of the examples. Specifically, this paper makes the following contributions: 

\begin{itemize}
    \item We present EXnet, a simple and scalable model architecture trained specifically to learn classification tasks with few-shot examples, without any parameter updates. 
    \item We evaluate EXnet on a collection of 9 unseen datasets (during training), which test both the cross-domain and cross-task understanding. Even the smallest variant of EXnet with 15 million parameters and only two in-context examples, shows performance gains of up to 48\% compared to GPT-Neo (1.3 billion parameters) which is 86$\times$ the size of our model.
\end{itemize}

\section{Engines}

\section{Related Work}
This section provides a comprehensive overview of the previous works related to EXnet. The overall landscape of few-shot learning can be grouped into two categories, few-shot learning with or without parameter updates. Generally, few-shot learning with parameter updates is termed as few-shot fine-tuning because it involves utilizing a small set of $K$ labeled samples to fine-tune a model. Gupta \etal \cite{gupta2020effective} explored transfer learning with BERT for few-shot fine-tuning, where they use text-label pairs as input, and predict if the label belongs to the text. Their results show that this simple approach outperforms all previous baselines over the ARSC dataset, suggesting that utilizing both text-label pairs as input has the potential to demonstrate cross-domain understanding. Wei \etal \cite{wei2021few} presented a data augmentation strategy powered by curriculum learning and BERT-based triplet networks for few-shot fine-tuning, where they show a slight 3\% improvement in performance over previous approaches. Ohashi \etal \cite{ohashi2021distinct} presented an approach where they utilize semantic knowledge from labels to learn distinct representations specific to each label, which in turn enables them to get improved model performance in few-shot fine-tuning.

Recently, prompt learning has also been explored for few-shot learning where PLMs are provided a task instruction/prefix which helps in utilizing their pre-trained knowledge to understand the task instruction and thus improving the few-shot performance. Schick \etal \cite{schick2020exploiting} presented PET, a prompt-based semi-supervised method that shows strong few-shot performance. PET utilizes cloze-style input patterns similar to Masked-Language modeling (MLM). Utama \etal \cite{utama2021avoiding} demonstrate through their experiments that few-shot fine-tuning performs poorly compared to zero-shot counterpart, which proves that few-shot fine-tuning degrades PLMs performance, particularly in the case of these sentence-pair classification tasks. A possible explanation for this phenomenon is that the PLMs lose their pre-trained knowledge by catastrophic forgetting during the few-shot gradient updates, and learns lexical shortcuts instead of meaningful patterns. 
An extreme case of few-shot learning without parameter updates is zero-shot learning, where no examples of the task are required. Yin \etal \cite{yin2019benchmarking} made the first attempt at unifying datasets and evaluating the zero-shot performance of language models over them which were only fine-tuned on natural language inference (NLI) tasks. However, Ma \etal \cite{ma2021issues} demonstrated some major drawbacks in NLI-based zero-shot text classification. They show that there are negligible gains on some tasks with NLI-based fine-tuning compared to standard BERT pre-trained on next-sentence prediction task. Moreover, they show that these NLI-based models only rely on shallow lexical patterns, along with high variance and a lack of stability. Zhong \etal \cite{zhong2021adapting} explored the idea of utilizing prompts in question-answering format which involve the task of predicting if the label description belongs to the text or not. They collect a large set of 43 datasets comprising 441 labels, which they categorize into groups of similar datasets and only evaluate on those groups that are unseen during training. Their experiments show that larger PLMs like T5-Large show much better performance compared to the smaller ones. In-context learning is the most promising variation of few-shot learning without gradient updates, which involves giving a set of label examples in the input to PLMs, along with a query text that must be predicted based on the understanding of the task from those examples. Brown \etal \cite{brown2020language} presented GPT-3, which was the first PLM to demonstrate this ability of In-context learning, mostly on text generation tasks. One of the major disadvantages of GPT-3 has been it's huge size (175 billion parameters), and it's range of biases that have been demonstrated by Zhao \etal \cite{zhao2021calibrate}. 

None of the prior work explored building a light-weight in-context model that can support a unlimited number of examples.                                           

\section{Proposed Model}
\begin{figure*}[ht]
    \centering
    \includegraphics[width=0.9\textwidth]{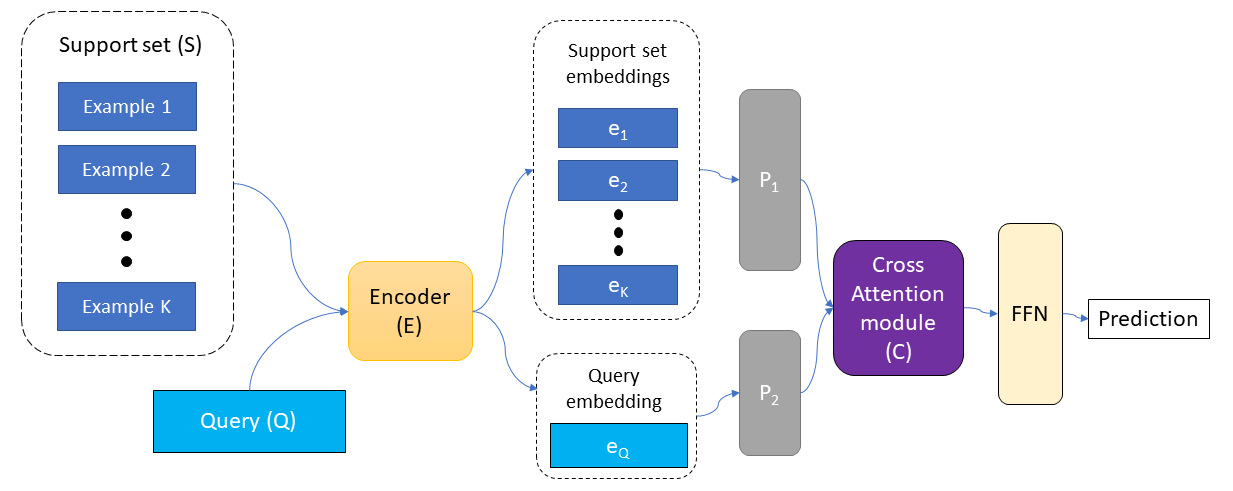}
    \caption{EXnet architecture}
    \label{fig:EXnet}
\end{figure*}
EXnet utilizes a Support set $S$ and a Query $Q$ as input which is a text in a question format with a fixed template $T$ as $T (x, l)$ = "question: is the text about <$l$>? [SEP] text: <$x$>". The support set $S$ consists of $K$ examples of text which belongs to the class $l$ mentioned in $Q$. Each support example also follows the format of $T$. For training, we use the Binary Cross-entropy loss function with the AdamW \cite{loshchilov2018decoupled} optimizer having a learning rate of 0.00002.
The architecture of EXnet has been schematically shown in Figure \ref{fig:EXnet}. The idea of using a single encoder to encode both the query and the support set, and then concatenating their embeddings for few-shot learning, was inspired by metric learning-based few-shot methods in Computer Vision, such as Relation Networks \cite{sung2018learning} and Matching Networks \cite{vinyals2016matching}.
EXnet consists of the following key components: 
\begin{itemize}
    \item \textbf{Encoder module: } We use a single encoder $E$ to generate embeddings for both the query $Q$ and support set $S$. Thus, $E$ generates $K$ embeddings $e_{1} ... e_{k}$ for $K$ examples in $S$ and one embedding $e_{q}$ for $Q$. The $K$ support set embeddings are then concatenated together into a single tensor $e_{S}$. Here, $E$ can be any encoder-only transformer model such as BERT \cite{devlin2018bert}. For simplicity, we experiment with three different variants of BERT based on size/number of parameters.
    \item  \textbf{Projector blocks: } Two projector blocks $P1$ and $P2$ are used to convert $e_{S}$ and $e_{Q}$ respectively into equal dimension representations. $P_{1}$ consists of three GeLU activated feed-forward layers with a 30\% dropout in between the second and third layer. $P_{2}$ consists of only a single GeLU activated feed-forward layer. We observe that keeping this asymmetry in projections enables the model to show slightly better performance.
    \item \textbf{Cross attention module: } The encoder $E$ only attends either query $Q$ or the support set $S$ at a time, without being able to utilize both of them at the same time. Thus, we use this cross-attention module to attend to both $Q$ and $S$ at the same time, which would enable EXnet to make use of the knowledge from the in-context examples in an effective manner to predict the answer to the query $Q$. The projected vectors are passed into this cross attention module $C$ which utilizes the first projected vector as Key and Value, and the second projected vector as Query in the Multi-Head Attention mechanism like the original Transformer\cite{vaswani2017attention}.
    \item \textbf{Feed-Forward network: } The output representation by the cross attention module is passed on to a Feed-Forward network (FFN) consisting of two GeLU activated Feed-Forward layers, followed by a final Sigmoid activated Feed-Forward layer which predicts the probability of the presence of class $l$ mentioned in $Q$.
\end{itemize}

\section{Experiments}
\subsection{Training setup}
We combined multiple public datasets representing the aspects of Topics, Emotions, Entities, and Questions, which play a key role in providing the essential amount of basic world knowledge to the models. To unify these datasets into the same binary ("yes"/"no") format, we use a simple strategy. For all the multi-class datasets, we keep all the labeled pairs of (text, class) as positives, and randomly sample an equal number of negatives for each text from other classes in the dataset. For every triplet (text, class, yes/no) in this prepared dataset, we sample a list of examples from all the positives. For training, we randomly sample input examples to optimize/calibrate the model to effectively use those examples. In constrast, during testing/evaluation we keep the set of examples fixed, which is further discussed in the next subsection.
Below, we describe about all the datasets used in this paper:
\begin{itemize}
    \item \textbf{GoEmotions: } It is a human-annotated dataset of 58k Reddit comments that were taken from well-known English-language subreddits and classified with 27 emotions\cite{demszky2020goemotions}. Most emotion classification datasets used in the realm of zero-shot/few-shot learning has been using the standard Ekman 6-way emotion categories. To provide much more fine-grained emotional knowledge of 27 emotions to the model, we chose GoEmotions.
    
    \item \textbf{News Category \footnote{\url{https://www.kaggle.com/datasets/rmisra/news-category-dataset}}: } This dataset contains around 200k news headlines and their respective categories (42) from the year 2012 to 2018 obtained from the news website called HuffPost. 

    \item \textbf{TREC-6 \cite{li-roth-2002-learning}: } This question classification dataset contains 5500 labeled questions in training set and another 500 for test set. It contains 6 labels comprising of questions about descriptions, abbreviations, entities, people/individuals, locations, and numerical/quantitative information.
    
    \item \textbf{OpenEntity \cite{choi2018ultra}: } This dataset consists of around 6,000 sentences that have been annotated with precise entity types. The target entity's appropriate types are described in the entity types, which are free-form noun phrases. Each sentence includes five categories on average. 
\end{itemize}

\subsection{Evaluation setup} \label{eval-setup}
We collected multiple public datasets on text classification to effectively evaluate our models. Similar to our training setup, we first convert every evaluation dataset into binary ("yes"/"no") format by the same strategy of sampling random negatives. To get examples, we create a set of examples for each class and use them for each and every text in that class. Unlike the training setup, we don't sample randomly from the entire dataset to make the evaluation robust and simulate a real-world scenario where humans using our model would be able to give only a few examples per label. We categorize these evaluation datasets into two different categories as described further in this section.

\subsubsection{Cross-domain evaluation:}
The purpose of cross-domain evaluation is to measure the model's capability to generalize across texts from different domains. For instance, \textbf{Abstract Classification} is similar to \textbf{News Category} as they both entail detecting topics from text but domains are different i.e., Scientific articles vs News. We gather the following datasets for this evaluation setup - 
\begin{itemize}
    \item \textbf{Cloth reviews\footnote{\url{https://www.kaggle.com/nicapotato/womens-ecommerce-clothing-reviews}}: } This is a binary classification dataset based on customer reviews of clothing items on an e-commerce website. The positive class entails whether the user recommends the item. The training task closest to this is the Emotion classification task i.e GoEmotions because predicting if a user recommends an item is almost the same as predicting a positive sentiment. Hence, this dataset will test if the model trained on Emotions is able to generalize to the domain of user reviews of clothing items, without being explicitly trained on that domain. 
    \item \textbf{Abstract classification\footnote{\url{https://www.kaggle.com/abisheksudarshan/topic-modeling-for-research-articles?select=Train.csv}}: } This is a multi-class topic classification dataset where each text is an abstract of a scientific article. The training task closest to this is News Category classification. This will test the model's capability to generalize on topic classification in the unseen domain of scientific text.
    \item \textbf{Water problem\footnote{\url{https://www.kaggle.com/vbmokin/nlp-reports-news-classification?select=water_problem_nlp_en_for_Kaggle_100.csv}}: } This is a multi-class topic classification dataset on the task of distinguishing text reports about water problems. It will evaluate the model's capability to generalize on the domain of water problems by understanding the chemical, biological, and other specialized concepts which were unseen during training.
    \item \textbf{Stock sentiment classification\footnote{\url{https://www.kaggle.com/yash612/stockmarket-sentiment-dataset}}: } This is a binary classification dataset for the task of identifying positive or negative trend in stock market by understanding user comments. Evaluating our model over this dataset would show if the model is capable of understanding/generalizing over domain-specific textual data about the stock market.
\end{itemize}

\subsubsection{Cross-task evaluation:}
The purpose of cross-task evaluation is to measure the model's capability to generalize across tasks that are unseen during training. We gather the following datasets for this evaluation setup - 
\begin{itemize}
    \item \textbf{ETHICS \cite{hendrycks2020aligning}: } This is a multi-class dataset with texts paired with labels of human moral values like  justice, well-being, duties, virtues, and commonsense morality. 
    \item \textbf{Irony detection: } This is a binary classification dataset with pairs of texts and corresponding labels representing if that text is an irony or not.
    \item \textbf{SMS spam detection: } This is a binary classification dataset comprising text and label pairs, where the positive class represents whether an SMS text message is spam and the negative class reflects if the SMS text message is not spam.
    \item \textbf{Stance classification: } This dataset is for the task of classifying the stance of an user over a particular topic (abortion, atheism, climate change, feminism, hillary) from his/her tweet's text.
    \item \textbf{SUBJ \cite{pang2004sentimental}: } This is a binary classification dataset with texts labeled with either being subjective or objective.
\end{itemize}

\section{Results and Discussion}
\begin{table*}[h!]
\centering
  \caption{Cross-domain evaluation}
  \label{tab:x-domain}
  \begin{tabular}{lcT{0.1\textwidth}T{0.1\textwidth}T{0.1\textwidth}T{0.1\textwidth}T{0.1\textwidth}}
   \toprule
    Dataset&K&EXnet-S&EXnet-M&EXnet-L&GPT-Neo&Zhong \etal\\
    \midrule
     & 2& 0.88& 0.92& 0.93&0.59&\\
    Cloth & 4& 0.88& 0.94& 0.93&0.63&-\\
     & 8& 0.89& 0.94& 0.95&0.62&\\
    \hline
     & 2& 0.69& 0.73&0.81&0.45&\\
    Abstract & 4& 0.70& 0.73&0.84&0.48&0.81*\\
     & 8& 0.71& 0.75&0.84&0.46&\\
    \hline
     & 2& 0.70& 0.72& 0.76&0.34&\\
    Water & 4& 0.71& 0.72& 0.76&0.37&-\\
     & 8& 0.71& 0.72& 0.76&0.38&\\
    \hline
     & 2& 0.67& 0.71& 0.72&0.30&\\
    Stock & 4& 0.70& 0.77& 0.76&0.30&-\\
     & 8& 0.74& 0.79& 0.80&0.28&\\
    \bottomrule
\end{tabular}
\end{table*}

\begin{table*}[h!]
\centering
  \caption{Cross-task evaluation}
  \label{tab:x-task}
  \begin{tabular}{lcT{0.1\textwidth}T{0.1\textwidth}T{0.1\textwidth}T{0.1\textwidth}T{0.1\textwidth}}
    \toprule
    Dataset&K&EXnet-S&EXnet-M&EXnet-L&GPT-Neo&Zhong \etal\\
    \midrule
     & 2& 0.48& 0.62&0.65&0.43&\\
    ETHICS & 4& 0.66& 0.70&0.74&0.44&-\\
     & 8& 0.73& 0.78&0.79&0.48&\\
    \hline
     & 2& 0.65& 0.68& 0.69&0.21&\\
    Irony & 4& 0.66& 0.72& 0.74&0.21&0.82*\\
     & 8& 0.69& 0.72& 0.75&0.28&\\
    \hline
     & 2& 0.24& 0.32& 0.34&0.14&\\
    SMS-spam & 4& 0.26& 0.33&0.37&0.18&0.35*\\
     & 8& 0.34& 0.39&0.42&0.22&\\
    \hline
     & 2& 0.78& 0.81&0.83&0.30&\\
    Stance & 4& 0.80& 0.85&0.87&0.36&0.74*\\
     & 8& 0.81& 0.84& 0.88&0.32&\\
    \hline
     & 2& 0.38& 0.43&0.48&0.39&\\
    SUBJ & 4& 0.46& 0.49&0.56&0.36&0.59*\\
     & 8& 0.67& 0.71&0.74&0.40&\\
    \bottomrule
\end{tabular}
\end{table*}

We use our evaluation datasets as discussed in Section \ref{eval-setup}, and measure EXnet's performance over each of them using the most commonly used classification metric, i.e, F1-score. We tend to use F1-score instead of accuracy because some of the evaluation datasets are imbalanced. We compare three sizes of EXnet which we call EXnet-S, EXnet-M, and EXnet-L based on the increasing order of model size. EXnet-S is a model with 15 million parameters which consists of Small BERT \cite{turc2019well} as the Encoder which has 6 layers and an embedding dimension of 256. EXnet-M is a 112 million parameter model which comprises of BERT-base as the encoder which has 12 layers and an embedding dimension of 768. EXnet-L is a 338 million parameter model that consists of BERT-large as the encoder which has 24 layers and an embedding dimension of 1024. As a baseline, we use the 1.3 billion parameter GPT-Neo model in an In-context setting (without fine-tuning) for comparing with EXnet. Unlike EXnet which requires only $K$ positive examples, GPT-Neo required both $K$ positives and $K$ negative examples as showing only positives leads it to predict only "yes"  for every query, previous work have shown that this is because GPT-Neo is trained on next word prediction. Furthermore, we also compare EXnet's performance with the meta-tuned model from Zhong \etal \cite{zhong2021adapting}.*Note: We report results of Zhong \etal from their original meta-tuned zero-shot (K=0) model. The results of our experiments are detailed in Table \ref{tab:x-domain} and Table \ref{tab:x-task}. It can be observed that EXnet is able to use the knowledge from the $K$ support examples, as increasing $K$ shows improved performance over most of the datasets. EXnet-M, and EXnet-L outperforms GPT-Neo over all the datasets in both cross-domain and cross-task evaluation by a large margin (up to 48\% gains in terms of F1 score). EXnet-S also outperforms GPT-Neo on all the datasets, except for SUBJ where GPT-Neo shows a slightly higher performance by a margin of 0.01 F1 score, when $K = 2$. In comparison to Zhong \etal, EXnet with larger $K$ is able to outperform their Meta-tuned model (770M parameters) over most of the evaluation datasets.
Based on our results, we make following observations -
\begin{itemize}
    \item \textbf{Benefits of increasing $K$ are dependent on task complexity: } In our cross-task evaluation, it can be seen from Table \ref{tab:x-task} that the complex tasks like SUBJ and ETHICS which involve very subjective human concepts, EXnet has a huge improvement when K is increased from 2 to 8 examples. This demonstrates that In-context learning has more utility whenever the concept required to understand the task is complex and subjective. 
    \item \textbf{In-context learning is more useful in improving cross-task performance than cross-domain: }
    As seen in \ref{tab:x-domain}, all the three variants of EXnet have the ability to understand cross-domain tasks, but increasing $K$ doesn't provide reasonable boost in performance. A possible reason for this is the fact that cross-domain evaluation involves tasks which were similar to training tasks, due to which EXnet was able to understand the task well with limited examples ($K=2$).
\end{itemize}

\section{Conclusion}
Zero-shot models do not provide sufficient accuracy for most real-world applications. We argue that In-context learning without gradient updates is better suited for most real-world applications. For example, to detect new set emotions in product reviews, one can take a model trained on similar task on publicly available datasets. Existing in-context learning models are huge and does not provide great task accuracy. 

In this paper, we presented EXnet, a lightweight model architecture specifically designed for In-context learning, that doesn't have limitation over number of examples. We evaluated EXnet  using 9 unseen datasets and demonstrated the strong cross-domain and cross-task understanding ability of EXnet, which even outperforms models that are 86$\times$ it's size. In the future, an interesting area to explore can be pre-training the encoder in EXnet with different strategies to observe the trends in performance. 

\bibliography{anthology,custom}
\bibliographystyle{acl_natbib}

\end{document}